\pdfoutput=1

\documentclass[11pt]{article}

\usepackage[]{emnlp2021}

\usepackage{times}
\usepackage{latexsym}

\usepackage[T1]{fontenc}

\usepackage[utf8]{inputenc}

\usepackage{microtype}

\usepackage{algorithm,algcompatible,amsmath}
\usepackage{graphicx}
\usepackage{url}
\usepackage{multirow}
\usepackage{comment}
\usepackage{amssymb}
\usepackage{pifont}
\usepackage{mathtools}
\usepackage{xcolor}
\usepackage{enumitem}
\usepackage{marvosym}
\usepackage{textcomp}
\usepackage{array}
\usepackage{booktabs}
\usepackage{multicol}
\usepackage{flushend}
\usepackage{amsfonts}
\usepackage{amsmath}
\usepackage{xcolor}
\usepackage{calc}
\usepackage{amssymb}
\usepackage{diagbox}
\usepackage{algorithm}  
\usepackage{algorithmicx}  
\usepackage{algpseudocode} 
\usepackage{ulem}
\usepackage{color}
\usepackage{array}
\makeatletter
\newcommand{\thickhline}{%
    \noalign {\ifnum 0=`}\fi \hrule height 1pt
    \futurelet \reserved@a \@xhline
}

\newcommand{\squishend}{
\end{list} }
\newcommand*\circled[1]{\kern-2.5em%
  \put(0,4){\color{white}\circle*{18}}\put(0,4){\circle{10}}%
  \put(-3,0){\color{black}\bfseries#1}~~}
\usepackage{tikz} 

\usepackage{enumitem}

\newcommand{\squishlist}{
\begin{list}{$\bullet$}
{ \usecounter{Lcount}
\setlength{\itemsep}{0pt}
\setlength{\parsep}{0pt}
\setlength{\topsep}{0pt}
\setlength{\partopsep}{0pt}
\setlength{\leftmargin}{2em}
\setlength{\labelwidth}{1.5em}
\setlength{\labelsep}{0.5em} } }
\newcommand{\modelname}{\texttt{GradLRE}}
\newcommand{\encoder}{Relational Label Generator}
\newcommand{\gradient}{Gradient Imitation Reinforcement Learning}
\newcommand{\augmentation}{Contextualized Data Augmentation}

\title{Gradient Imitation Reinforcement Learning \\ for Low Resource Relation Extraction}

\author{Xuming Hu$^1$, Chenwei Zhang$^{2\dagger}$, Yawen Yang$^1$, Xiaohe Li$^1$, Li Lin$^1$,\\\textbf{Lijie Wen$^{1\dagger}$, Philip S. Yu$^{1,3}$}\\
  $^1$Tsinghua University,  $^2$Amazon, $^3$University of Illinois at Chicago\\
  $^1$\texttt{\{hxm19,yyw19,lixh18,lin-l16\}@mails.tsinghua.edu.cn}\\
  $^1$\texttt{wenlj@tsinghua.edu.cn}
  $^2$\texttt{cwzhang@amazon.com}  
  $^3$\texttt{psyu@uic.edu}\\
  }

\begin{document}
\maketitle

\begin{abstract}

Low-resource Relation Extraction (LRE) aims to extract relation facts from limited labeled corpora when human annotation is scarce. Existing works either utilize self-training scheme to generate pseudo labels that will cause the gradual drift problem, or leverage meta-learning scheme which does not solicit feedback explicitly. To alleviate selection bias due to the lack of feedback loops in existing LRE learning paradigms, we developed a {\gradient} method to encourage pseudo label data to imitate the gradient descent direction on labeled data and bootstrap its optimization capability through trial and error. We also propose a framework called {\modelname}, which handles two major scenarios in low-resource relation extraction. Besides the scenario where unlabeled data is sufficient, {\modelname} handles the situation where no unlabeled data is available, by exploiting a contextualized augmentation method to generate data. Experimental results on two public datasets demonstrate the effectiveness of {\modelname} on low resource relation extraction when comparing with baselines. Source code is available\footnote{\url{https://github.com/THU-BPM/GradLRE}\\\phantom{00} $^\dagger$Corresponding Authors.}.

\end{abstract}
\section{Introduction}\label{introduction}
Relation Extraction (RE) aims to discover the semantic relation that holds between two entities and transforms massive corpus into structured triplets (entity$_{head}$, relation, entity$_{tail}$). For example, from ``A \textit{letter}$_{head}$ was delivered to my \textit{office}$_{tail}$...", we can extract a relation ${\texttt{Entity-Destination}}$ between head and tail entities.
Neural RE methods leverage high-quality annotated data or human curated knowledge bases to achieve decent results \citep{zeng2017incorporating,zhang2017position}. However, these manually labeled data would be labor-intensive to obtain. This motivates a Low Resource Relation Extraction (LRE) task where annotations are scarce.

Lots of efforts are devoted to improve the model generalization ability beyond learning directly from existing, limited annotations. Distant Supervision methods leverage facts stored in external knowledge bases (KBs) to obtain annotated triplets as the supervision \citep{mintz2009distant,zeng2015distant}. However, these methods should make a strong assumption that two co-occurring entities convey KB relations regardless of specific contexts, which makes model generate relations based on contextless rules and limits the generalization ability. To leverage unlabeled data, \citet{rosenberg2005semi} propose to assign pseudo labels on unlabeled data and leverage pseudo labels to iteratively improve the generalization capability of the model. However, during the training process, self-training models suffer from the
gradual drift problem \citep{curran2007minimising,zhang2016understanding} caused by noisy pseudo labels. \citet{hu2020semi} alleviate the noise in pseudo labels by adopting a meta-learning scheme during pseudo label generation, then leveraging pseudo label selection and exploitation scheme to obtain high-confidence pseudo labels. However, when limited annotations are directly used during training, the trained models inevitably possesses selection bias towards, if not overfit on, limited labeled data, which impedes LRE models from further generalizing beyond the annotations.

\begin{figure}[bt!]
    \centering
    \includegraphics[width=0.7\linewidth]{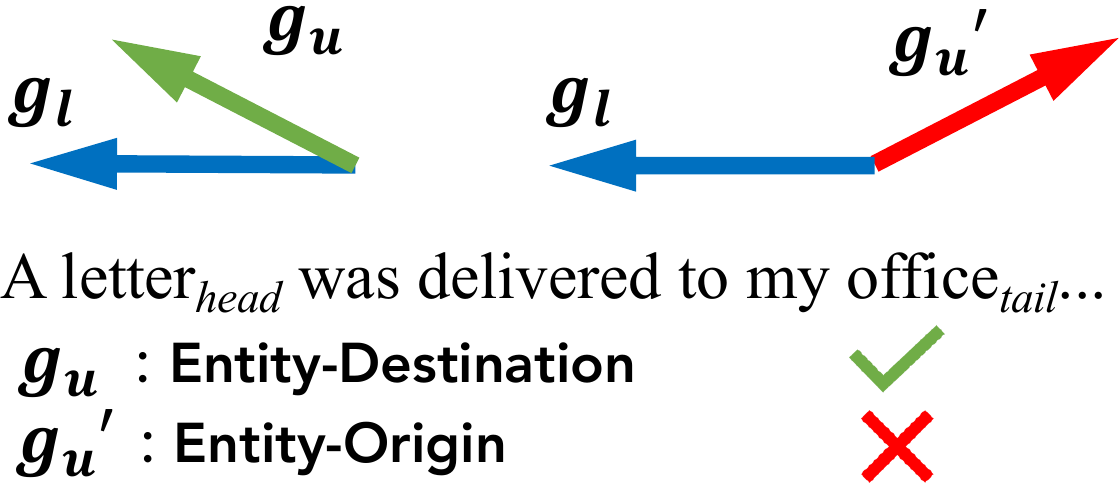}
    \caption{Gradient descent direction on labeled data (${g_{l}}$) and unlabeled data with correct or incorrect pseudo label (${g_{u}}$, ${g_{u}^{'}} $).}
    \label{fig:introduction}
\vspace{-0.1in}
\end{figure}
To improve the generalization ability for LRE,
we propose to use existing annotations as a guideline instead of having them directly involved in training, as well as introducing an explicit feedback loop when consuming annotations.
More specifically, we first encourage pseudo-labeled data to imitate labeled data on the gradient descent directions during the optimization process.
We illustrate this idea in Figure \ref{fig:introduction}.
${g_{l}}$ represents the average gradient descent direction on labeled data. ${g_{u}}$ and ${g_{u}^{'}}$ represent the correct and incorrect pseudo labels on unlabeled data, which guides the gradient descent direction in a positive/negative fashion \cite{du2018adapting,sariyildiz2019gradient,yu2020gradient}. 
Based on how well the pseudo-labeled data mimics the instructive gradient descent direction obtained from limited labeled data, we then design a reward to quantify the behavior and aim to use the reward as an explicit feedback.
This learnable setting can be naturally formulated into a reinforcement learning framework, which aims to learn an imitation policy that maximizes the reward through trial and error. When comparing with methods where annotations are directly used in the traditional learning schema, this formulation also allows a feedback mechanism and thus increases generalization ability beyond limited annotations. We name our method as {\gradient} in this paper.

We propose a framework called {\modelname}, which integrates {\gradient} and is able to handle two major scenarios in LRE: 
1) a typical scenario when limited labeled data and large amounts of unlabeled data are available, and an extreme yet practical scenario where 2) even unlabeled data is absent: only limited labeled data is available. {\modelname} handles the former scenario via pseudo labeling optimized through {\gradient} and tackles the later scenario by using a {\augmentation} module.

To summarize, the main contributions of this work are as follows:
\begin{itemize}
\item We propose a gradient imitation reinforcement learning method that alleviates the bias from training directly with limited annotation, and encourages the RE model to effectively generalize beyond limited annotations.
\item We develop a LRE framework {\modelname} that handles two low-resource relation extraction scenarios by leveraging both {\gradient} and {\augmentation}.
\item We show that {\modelname} outperforms strong baselines. Extensive experiments validate the effectiveness of the proposed techniques.
\end{itemize}

\section{Proposed Model}
\begin{figure}[bt!]
    \centering
    \includegraphics[width=\linewidth]{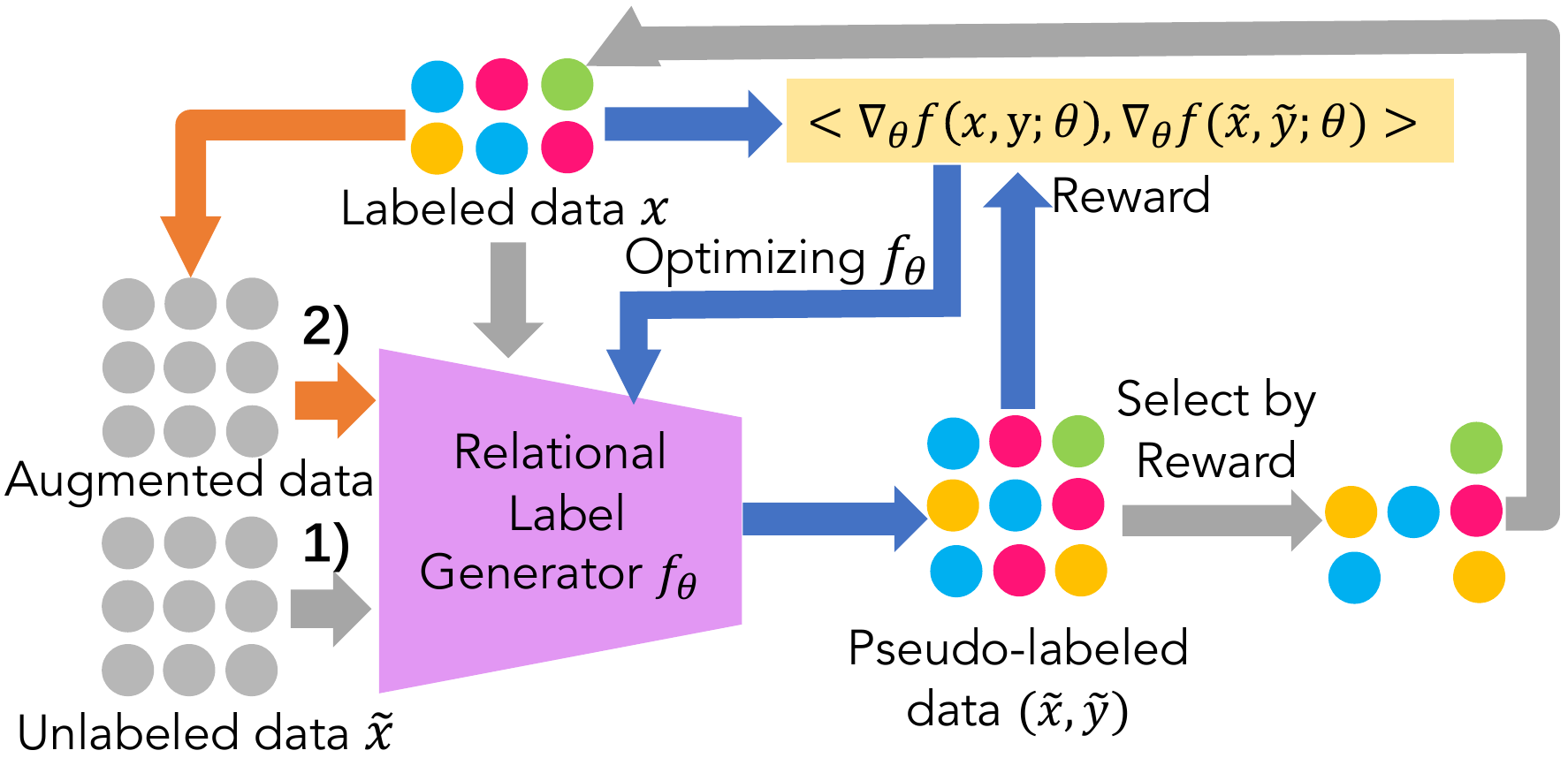}
    \caption{Overview of the proposed {\modelname} framework for Low Resource Relation Extraction. 1) and 2) represent two LRE scenarios, respectively. {\color{blue}{Blue}} arrows represent Gradient Imitation Reinforcement Learning and {\color{orange}{Orange}} represents Contextualized Data Augmentation. }
    \label{fig:overview}
\vspace{-0.1in}
\end{figure}

The proposed framework {\modelname} consists of three modules: {\encoder} (RLG), {\gradient} (GIRL) and {\augmentation} (CDA). As illustrated in Figure \ref{fig:overview}, two low resource relation extraction scenarios are handled. For the first scenario where limited labeled data and large amounts of unlabeled data are available, the input of RLG is labeled data and unlabeled data. Labeled data consists of sentences and relation mentions: [Sentence, Entity${_{1}}$, Entity${_{2}}$, Relation]. For the second scenario where only limited labeled data is available, we adopt CDA to generate unlabeled data and utilize these unlabeled data the same way as in the first scenario.

In a traditional self-training setting, we fine-tune RLG directly using the labeled data, and let RLG assign pseudo labels on unlabeled data as pseudo-labeled data. However, we argue that such learning paradigm suffers from selection bias due to the lack of feedback loops: the bias occurs when a model itself influences the generation of data which is later used for training.
In this work, we complete the feedback loop and alleviate such bias by leveraging GIRL to learn a policy that maximizes the likelihood between the expected gradient optimization direction from pseudo labels, and the average gradient optimization direction on labeled data.

\subsection{\encoder}
The {\encoder} (RLG) aims to obtain contextualized relational features for each input sentence based on the entity pair, and classify the entity pair into specific relations. In this work, we assume named entities in the sentence have been recognized in advance.

For a sequence of words in a sentence ${x}$ where two entities $E1$ and $E2$ are mentioned, we follow the labeling schema adopted in \citet{soares2019matching} and argument ${x}$ with four reserved tokens to mark the beginning and the end of each entity. We inject the ${[E1]}$, ${[/E1]}$, ${[E2]}$, ${[/E2]}$ to $x$
as the input token sequence for RLG, for example, ``A ${[E1]}$ letter ${[/E1]}$ was delivered to my ${[E2]}$ office ${[/E2]}$...". Considering that the relational representation between entity pairs are usually contained in the context, we leverage pretrained deep bi-directional transformers networks: BERT \citep{devlin2019bert} to effectively encode entity pairs, along with their context information. We concatenate the outputs corresponding to ${[E1]}$ , ${[E2]}$ positions as
$\mathbf{h}\in\mathbb{R}^{2\cdot{h_{R}}}$ where ${\mathbf{h} = [\mathbf{h}_{[E1]}, \mathbf{h}_{[E2]}]}$ and ${h_{R}}$ is the contextualized relational representation length. The RLG then classifies these representations into specific relations with a fully connected network. We adopt this architecture to generate labels on sentences, and denoted the RLG process as $f_{\theta}(x,E1,E2)$.

\subsection{\gradient}
Generally, we assign pseudo labels via RLG on unlabeled data as pseudo-labeled data, and add the selected pseudo-labeled data into the existing labeled data to iteratively improve RLG. 
We argue that without a feedback loop measuring the quality of pseudo labels, the model is more likely to suffer from selection bias and is impeded towards a better generalization ability.

We aim to generate pseudo labels with less labeling biases and errors especially with scarce annotations. 
To achieve this goal, we focus on improving the RLG performance by introducing gradient imitation to define and quantify what an appealing behavior looks like. We define the partial derivatives of the loss function corresponding to RLG parameters on the labeled data as standard gradient descending, and assume that when pseudo-labeled data are correctly labeled in RLG, partial derivatives to the RLG parameters on the pseudo-labeled data would be highly similar to standard gradient descending. Following this assumption, we propose Gradient Imitation Reinforcement Learning (GIRL), which optimizes RLG under a reinforcement learning framework \cite{williams1992simple}. Now we explain the reinforcement learning process in detail.

\noindent\textbf{State}: 
State is used to signal the optimization status. We use ${s^{(t)}}$ to denote the state. ${s^{(t)}}$ consists of the updated labeled dataset ${\mathcal{D}_{l}}$ at step $t$, along with a standard gradient direction $g_{l}$ at step $t$.

\noindent\textbf{Policy}: Our policy is learned to assign correct pseudo label on unlabeled data. The policy network is parameterized by the RLG network $f_{\theta}$.

\noindent\textbf{Action}: The action is to predict relational label on unlabeled data ${\widetilde{x}^{(t)}}$ as pseudo-labeled data ${(\widetilde{x}^{(t)},\widetilde{y}^{(t)})}$ given the State at step $t$. We consider the relation that corresponds to the maximum probability after softmax as the pseudo label:
\begin{equation}
\widetilde{y}^{(t)}=\operatorname{argmax}(f_{\theta}(\widetilde{x}^{(t, E1, E2)})).
\end{equation}
\noindent\textbf{Reward}: We use reward to signal labeling biases from the current policy on pseudo-labeled data. Our goal is to minimize the approximation error of the gradients obtained over the pseudo-labeled data. In other words, we maximize the correlation between gradients over the pseudo-labeled data and those over the labeled data. 

We define the standard gradient descent direction on the all ${N}$ labeled data as ${g_{l}}$ and the expected gradient descent direction on the pseudo-labeled data as ${g_{p}}$ respectively:
\begin{align}
&{g_{l}}^{(n)}(\theta)=\nabla_{\theta} \mathcal{L}_{l}\left(x^{(n)},y^{(n)} ; \theta\right),\\
&{g_{p}}^{(t)}(\theta)={\nabla_{\theta}} \mathcal{L}_{p}\left( \widetilde{x}^{(t)},\widetilde{y}^{(t)}; \theta\right),
\end{align}
where 
${\nabla_{\theta}}$ refers to the partial derivatives of the cross entropy loss $\mathcal{L}$ corresponding to Policy ${f_{\theta}}$ with respect to ${{\theta}}$. 
Considering that the outliers in the labeled data will affect the direction of standard gradient descent, we approximate ${g_{l}}$ over all ${N}$ labeled data and we define $\mathcal{L}_{l}$ and $\mathcal{L}_{p}$ as:
\begin{align}
\mathcal{L}_{l} =& \frac{1}{N}\sum_{n=1}^{N}\mathit{loss}(f_{\theta}({x}^{(n, E1, E2)}),\operatorname{one\_hot}(y^{(n)})),\\
\label{generated_loss}\mathcal{L}_{p} =& \mathit{loss}(f_{\theta}(\widetilde{x}^{(t, E1, E2)}),\operatorname{one\_hot}(\widetilde{y}^{(t)})),
\end{align}
where $\mathit{loss}$ is the cross entropy loss function, ${f_{\theta}({x}^{(n, E1, E2)})}$ returns a probability distribution over all relation categories for the $n$-th sample and ${\operatorname{one\_hot}(y^{(n)})}$ returns a one-hot vector indicating the target label assignment.

Since the most important guidance obtained by the gradient vector ${g_{l}}$ is its gradient descending direction, so we measure the discrepancy between ${g_{l}}$ and ${g_{p}}$ for state ${s^{(t)}}$ by defining their cosine similarity 
as the reward:
\begin{align}
R^{(t)}=\frac{g_{l}(\theta)^{\mathrm{T}} g_{p}(\theta)}{\left\|g_{l}(\theta)\right\|_{2}\left\|g_{p}(\theta)\right\|_{2}}.
\end{align}
The range of ${R^{(t)}}$ is [-1,1]. For those pseudo-labeled data ${(\widetilde{x}^{(t)},\widetilde{y}^{(t)})\in\mathcal{D}_{p}}$ with ${R^{(t)}>\lambda}$, ${\lambda=0.5}$, we treat them as positive reinforcement to improve the generalization ability of RLG network. We add these selected pseudo-labeled data to the labeled data and correct the standard gradient descending direction:
\begin{align}
&\mathcal{D}_{l}\leftarrow \mathcal{D}_{l}\cup\mathcal{D}_{p},\\
\label{eq:standard gradient} g_{l}\leftarrow& \frac{1}{N+1}(Ng_{l}+g_{p}).
\end{align}
For Eq. \eqref{eq:standard gradient}, we set the weight of the updated gradient direction according to the number of samples, where the standard gradient direction is calculated using all $N$ labeled samples and each pseudo labeled sample. The positive feedback obtained from GIRL via trial and error can attribute the improvement of RLG network (Policy) to assign correct pseudo label for next unlabeled data ${\widetilde{x}^{(t)}}$ (State). 

\noindent\textbf{Reinforcement Learning Loss}\\
We adopt the REINFORCE algorithm \cite{williams1992simple} and Policy Gradient for optimization. We calculate the loss over a batch of pseudo-labeled samples. The RLG will be optimized by GIRL on each batch according to the following reinforcement learning loss:
\begin{align}
\begin{split}
\mathcal{L}(\theta) = \sum_{t=1}^{T} \mathit{loss} \big(f_\theta(\widetilde{x}^{(t, E1, E2)}),\\ \operatorname{one\_hot}(\widetilde{y}^{(t)})\big)*R^{(t)},
\end{split}
\end{align}
where $\mathit{loss}$ is the cross entropy loss function, ${R^{(t)}}$ is the reward and ${\widetilde{y}^{(t)}\sim\pi(\cdot|\widetilde{x}^{(t, E1, E2)};\theta)}$. The $\pi$ function means Policy in reinforcement learning. In our setting, it is parameterized as $f_{\theta}$, which is learned to assign pseudo labels on unlabeled data and we minimize ${\mathcal{L}(\theta)}$ to optimize the ${\theta}$. $T$ represents a total number of time steps in a reinforcement learning episode and is set to $16$, the same number as the batch size. For each high reward ${R^{(t)}>\lambda}$, ${\lambda=0.5}$ pseudo-labeled data, we use it to dynamically update the labeled dataset / standard gradient direction and guide the reinforcement learning process to the next State.

Note that ${f_{\theta}}$ is first pretrained using all the labeled data in a supervised way. During the process of calculating reinforcement learning loss, our model follows the Markov’s decision process and the labeled data ${\mathcal{D}_{l}}$ and standard gradient descending direction ${g_{l}}$ will be dynamically corrected by the selected pseudo-labeled data ${\mathcal{D}_{p}}$, which means that for each State, Policy will be updated over time $t$. The RLG could solicit positive feedback obtained using GIRL via trial and error.

\subsection{\augmentation}
Except the typical LRE scenario where both limited labeled data and large amounts of unlabeled data are available, {\modelname} handles an extreme yet practical LRE scenario additionally, where only limited labeled data is available. 
As shown by the orange arrow in Figure \ref{fig:overview},
we propose to use a contextualized augmentation method, namely CDA, to generate more unlabeled data.

Given a sentence ${x}$ where two entities $E1$ and $E2$ are mentioned in the labeled data, CDA samples spans of the sentence as \texttt{[MASK]} until the masking budget has been spent (e.g., 15\% of ${x}$) and finally fills the mask with tokens using the pretrained language model. Inspired by \citet{joshi2020spanbert}, we sample a span length from a geometric distribution ${\ell \sim \operatorname{Geo}(p)}$ where ${\ell \in [1,10]}$. ${p}$ will affect the probability of selecting different span lengths. A larger ${p}$ leads to a shorter span. We follow \citet{joshi2020spanbert} and choose ${p=0.2}$. The ${\operatorname{Geo}(0.2)}$ yields a mean span length of ${(\ell)=3.8}$ and shorter spans are more inclined to be chosen. 
We skip $E1$ and $E2$ as \texttt{[MASK]} and also require the starting point of the span must be the beginning of one word which ensures to mask complete words. 

For example, we may mask the word \textsc{delivered to} in ``A letter was delivered to my office in this morning.'' and obtain an augmented sentence ``A letter was sent from my office in this morning.''. 
Compared with the original labeled data, the augmented sentence may have a different relation label. We therefore use RLG, which has a strong discriminate power,
to assign a correct label to the augmented unlabeled sentence. Since ``no relation'' has been defined as one valid relation category in the dataset, RLG has the capability to safely assign one augmented sentence as ``no relation'' when it is out of scope.

\begin{table*}[th!]
\centering
  \resizebox{.95\linewidth}{!}{
\begin{tabular}{lcccccc}
\thickhline
\multicolumn{1}{l}{\multirow{2}{*}{\textbf{Methods / \%Labeled Data}}} & \multicolumn{3}{c}{\textbf{SemEval}} & \multicolumn{3}{c}{\textbf{TACRED}} \\ \cmidrule(lr){2-4}\cmidrule(lr){5-7}
\multicolumn{1}{c}{}                         & 5\%     & 10\%    & 30\%    & 3\%    & 10\%    & 15\%    \\ \hline
                           LSTM \citep{hochreiter1997long}             &  22.65\small±3.35        &   32.87\small±6.79       &   63.87\small±0.65       &  28.68\small±4.29       &  46.79\small±0.99       &   49.42\small±0.59       \\
                          PCNN \citep{zeng2015distant}              &  41.82\small±4.48       &    51.34\small±1.87     &  63.72\small±0.51       &  40.02\small±5.23       &  50.35\small±3.28    &  52.50\small±0.39   \\
                           PRNN \citep{zhang2017position}                  &  55.34\small±1.08    & 62.63\small±1.42  & 69.02\small±1.01  & 39.11\small±1.92  &  52.23\small±1.20  & 54.55\small±1.92   \\
                          BERT \citep{devlin2019bert}             &  70.71\small±1.24  &  71.93\small±0.99   &  78.55\small±0.87  & 40.11\small±3.88  & 53.17\small±1.67   &  55.55\small±0.82   \\\hline \hline
                       Self-Training$_{BERT}$ \citep{rosenberg2005semi}       & 71.34\small±1.68  &  74.25\small±1.10  & 81.71\small±0.79  & 42.11\small± 1.04    &  54.17\small±0.53  &   56.52\small±0.40 \\
                        Mean-Teacher$_{BERT}$ \citep{tarvainen2017mean}     & 70.05\small±3.89   & 73.37\small±1.42  & 80.61\small±0.81  &  44.34\small±1.78  & 53.08\small±1.01   & 53.79\small±1.38 \\
                      RE-Ensemble$_{BERT}$ \citep{lin2019learning}         & 72.35\small±2.63    &  75.71\small±1.39 & 81.34\small±0.74  &  42.78\small±1.89   &  54.83\small±0.95    & 55.68\small±1.21   \\
                      DualRE-Pairwise$_{BERT}$ \citep{lin2019learning}         & 74.35\small±1.76    &  77.13\small±1.10 & 82.88\small±0.67  &  43.06\small±1.73   &  56.03\small±0.55    & 57.99\small±0.67   \\
                      DualRE-Pointwise$_{BERT}$ \citep{lin2019learning}          & 74.02\small±1.68    &  77.11\small±1.02 & 82.91\small±0.62  &  43.73\small±1.60   &  56.28\small±0.61    & 57.72\small±0.49   \\
                     MRefG$_{BERT}$ \citep{li2020exploit}             &  75.48\small±1.34   & 77.96\small±0.90    & 83.24\small±0.71  &  43.81\small±1.44   &  55.42\small±1.40    &  58.21\small±0.71   \\
                    MetaSRE$_{BERT}$ \citep{hu2020semi}          &  78.33\small±0.92   & 80.09\small±0.78   &   84.81\small±0.44  &  46.16\small±1.02   & 56.95\small±0.34    &  58.94\small±0.36   \\
                     \textbf{{\modelname}$_{BERT}$ (Ours)}               &  \textbf{79.65\small±0.68}       &  \textbf{81.69\small±0.57}        & \textbf{85.52\small±0.34}   &  \textbf{47.37\small±0.74}  &  \textbf{58.20\small±0.33}    &  \textbf{59.93\small±0.31}    \\\hline \hline
                     BERT w. gold labels            &   84.64\small±0.28       & 85.40\small±0.34        &   87.08\small±0.23      &    62.93\small±0.41    &  63.66\small±0.23        &   64.69\small±0.29     \\\thickhline
\end{tabular}}
\caption{F1 (\%) comparisons on the SemEval and TACRED datasets with various amounts of labeled data and 50\% unlabeled data.}\label{tab:Result}
\vspace{-0.1in}
\end{table*}

\section{Experiments}
We conduct extensive experiments on two datasets to prove the effectiveness of our Gradient Imitation Reinforcement Learning for low resource relation extraction tasks, and give a detailed analysis of each module to show the advantages of {\modelname}.
\subsection{Datasets}
We follow \citet{hu2020semi} to conduct experiments on two public RE datasets, including the SemEval 2010 Task 8 (\textbf{SemEval}) \citep{hendrickx2010semeval}, and the TAC Relation Extraction Dataset (\textbf{TACRED}) \citep{zhang2017position}. SemEval is a standard benchmark dataset for evaluating relation extraction models, which consists of training, validation, test set with 7199, 800, 1864 relation mentions respectively, with 19 relations types in total (including \textit{no\_relation}), of which \textit{no\_relation} percentage is 17.4\%. TACRED is a large-scale crowd-sourced relation extraction dataset which is collected from all the prior TAC KBP relation schema. The dataset consists of training, validation, test set with 75049, 25763, 18659 relation mentions respectively, with 42 relation types in total (including \textit{no\_relation}), of which \textit{no\_relation} percentage is 78.7\%.

\subsection{Baselines and Evaluation metrics}
{\modelname} is flexible to integrate different contextualized encoders. From Table \ref{tab:Result}, we first compare several widely used supervised relation encoders with only labeled data: \textbf{LSTM} \citep{hochreiter1997long}, \textbf{PCNN} \citep{zeng2015distant}, \textbf{PRNN} \citep{zhang2017position}, \textbf{BERT} \citep{devlin2019bert}. Among them, BERT achieved the state-of-the-art performance. So we adopt BERT as the base encoder for both {\modelname} and other baselines for a fair comparison.

For baselines, we compare {\modelname} with other six representative methods:
(1) \textbf{Self-Training} \citep{rosenberg2005semi} iteratively improves model by predicting unlabeled data with pseudo labels and adds these pseudo label data to labeled data. 
(2) \textbf{Mean-Teacher} \citep{tarvainen2017mean} is jointly optimized by a perturbation-based loss and a training loss to ensure that the model makes consistent predictions on similar data. 
(3) \textbf{DualRE} \citep{lin2019learning} treats relation extraction as a dual task from relations to sentences and combines the loss of a prediction module and a sentence retrieval module. The difference between Pairwise and Pointwise schemes lie in whether the retrieved documents are given scores or a relative order. 
(4) \textbf{RE-Ensemble} \citep{lin2019learning} replaces the retrieval module in the proposed DualRE framework with the same prediction module. 
(5) \textbf{MRefG} \citep{li2020exploit} semantically connects the unlabeled data to the labeled data by constructing reference graphs, including entity reference, verb reference and semantics reference. 
(6) \textbf{MetaSRE} \citep{hu2020semi} is the state-of-the-art method that generates pseudo labels on unlabeled data by meta learning from the successful and failed attempts on classification module as an additional meta-objective.

Finally, we present another model: \textbf{BERT w. gold labels}, which indicates the upper bound of LRE models when all unlabeled data has gold labels during training with labeled data.

For the evaluation metrics, we choose F1 score as the main metric. Note that following \citet{hu2020semi}, the correct predictions of \textit{no\_relation} are ignored.
\subsection{Implementation Details}\label{implementation details} 
For the two datasets, strictly following the settings used in \citet{hu2020semi}, we use stratified sampling to divide training set into labeled and unlabeled datasets of various proportions to ensure all subsets share the same relation label distribution. For SemEval, we sample 5\%, 10\% and 30\% of the training set, for TACRED, we sample 3\%, 10\% and 15\% of the training set as labeled datasets. For both datasets, we sample 50\% of the training set as unlabeled dataset. As suggested in \citet{hu2020semi}, we split all unlabeled data into 10 segments. In each iteration, RLG is optimized based on one segment of the data. The RLG gradually improves as we obtain more high-quality pseudo labels one iteration after another. We implement this strategy for our model and the baselines. For the evaluation metrics, we choose F1 score as the main metric.

For RLG, we use the BERT default tokenizer with max-length as 128 to preprocess data. We use pretrained BERT-Base\_Cased as the initial parameter to encode contextualized entity-level representation. The fully connected network is defined with layer dimensions of ${2{h_{R}}}$-${h_{R}}$-label\_size, where ${h_{R}}=768$. We use BertAdam with $1e{-4}$ learning rate and warmup with 0.1 to optimize the loss. For GIRL, the total time step ${T}$ is set to 16, the same number as the batch size. We use AdamW \cite{loshchilov2018fixing} with $5e{-5}$ learning rate to optimize the reinforcement learning loss.

\subsection{Main Results}
Table \ref{tab:Result} shows the mean and standard deviation F1 results with 5 runs of training and testing on SemEval and TACRED when leveraging various labeled data and 50\% unlabeled data. All methods could gain performance improvements from the unlabeled data when compared with the model that only uses labeled data (BERT), which demonstrates the effectiveness of unlabeled data in the LRE setting. We could observe that {\modelname} outperforms all baseline models consistently. More specifically, compared with the previous SOTA model MetaSRE, {\modelname} on average achieves 1.21\% higher F1 on SemEval and 1.15\% higher F1 on TACRED across various labeled data. When considering standard deviation, {\modelname} is also more robust than all the baselines.

Considering LRE when labeled data is very scarce, e.g. 5\% for SemEval and 3\% for TACRED, {\modelname} could achieve an average 1.27\% F1 boost compared with MetaSRE. When more labeled data is available,  30\% for SemEval and 15\% for TACRED, the average F1 improvement is consistent, but reduced to 0.85\%. We attribute the consistent improvement of {\modelname} to the explicit feedback which GIRL is adopted and learning via trial and error: we use Gradient Imitation as a proxy for the classification loss in optimizing RLG.
The guidance from the gradient direction, as a part of the gradient imitation process, is more instructive, explicit, and generalizable than the implicit signals from training directly on labeled data.

\begin{figure}[t!]
    \centering
    \includegraphics[width=0.49\linewidth]{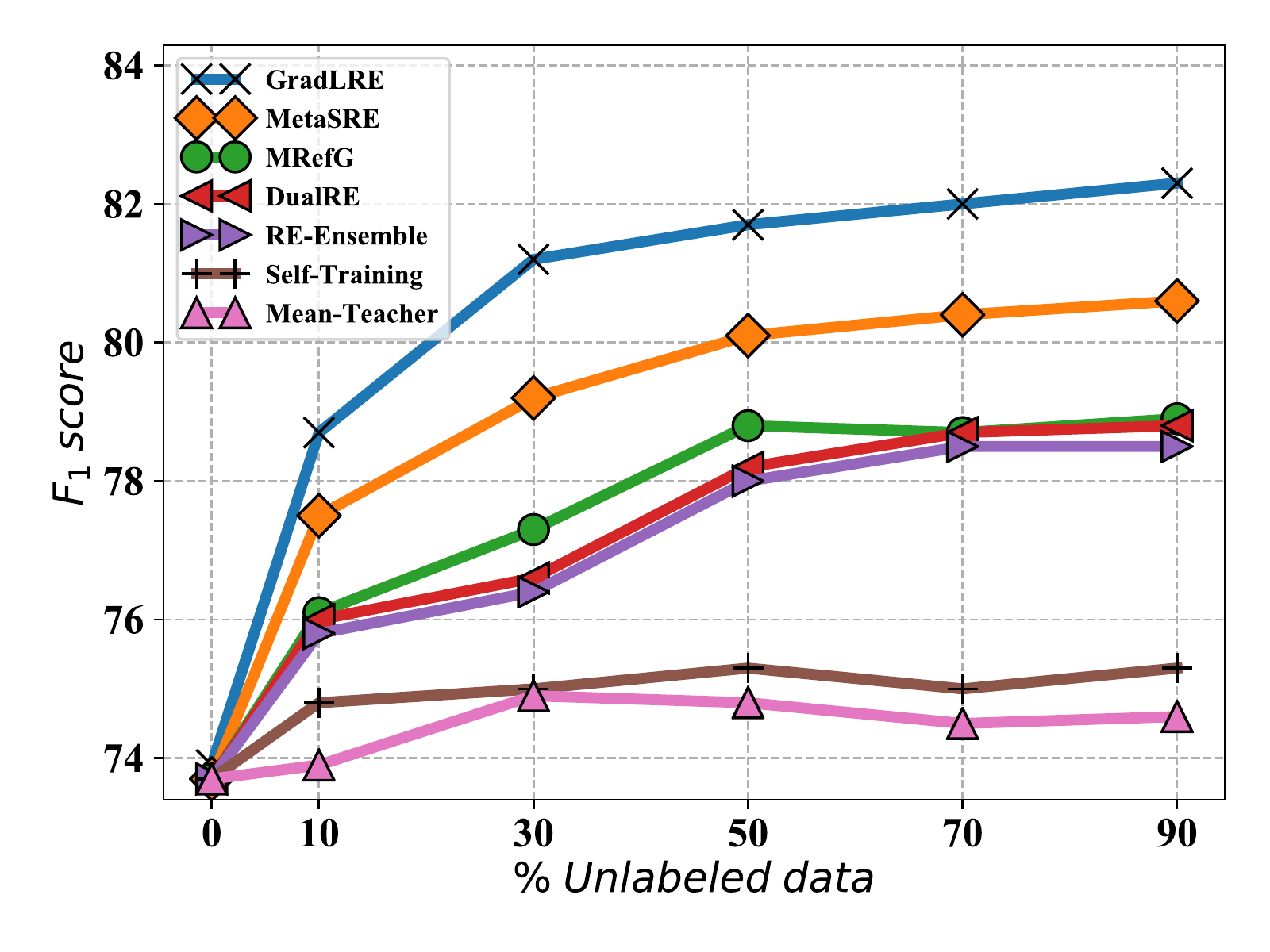}
    \includegraphics[width=0.49\linewidth]{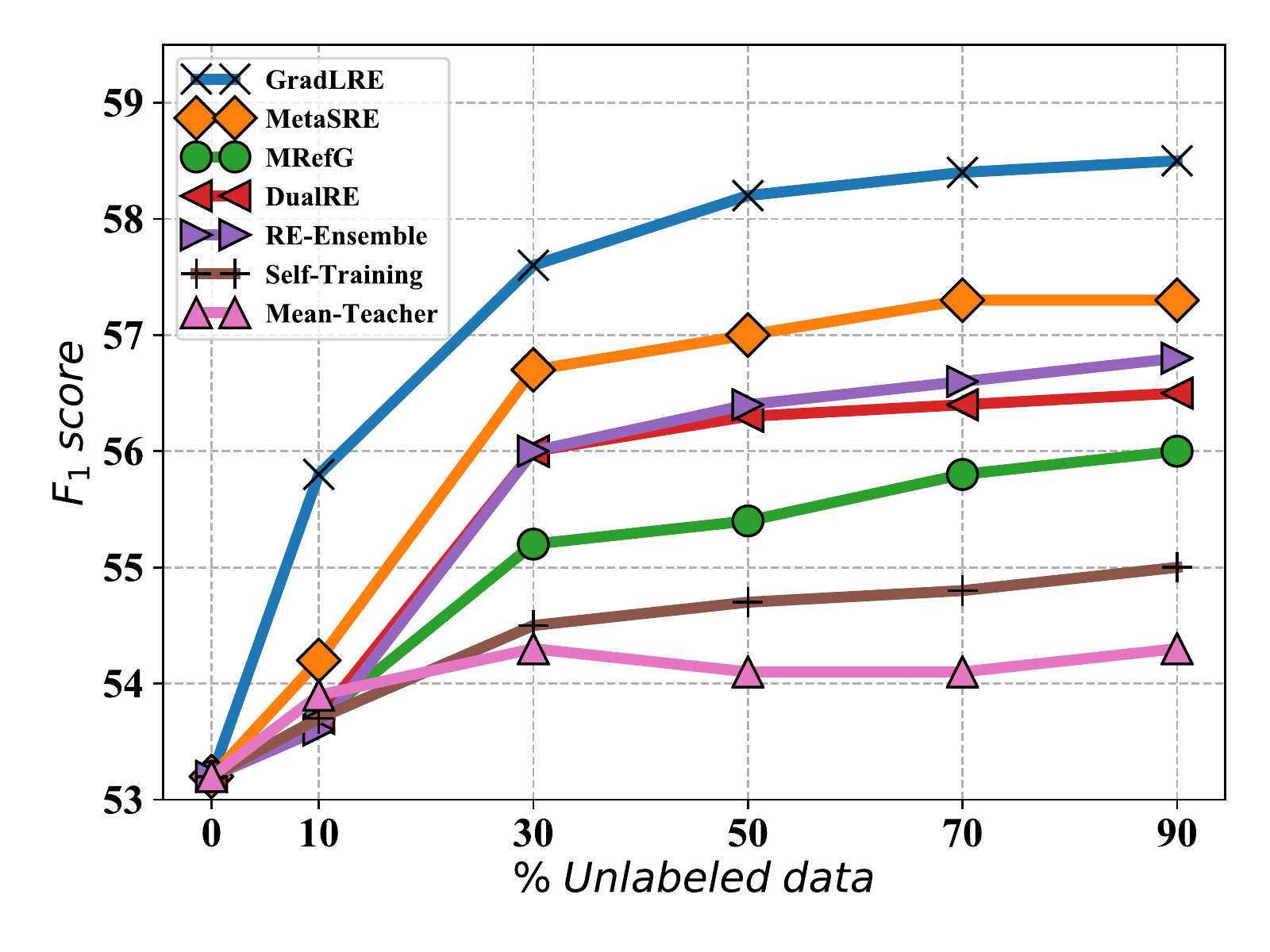}
    \caption{F1 (\%) Performance with various unlabeled data and 10\% labeled data on SemEval (left) and TACRED (right).}
    \label{fig:unlabel}
\vspace{-0.1in}
\end{figure}

We further vary the ratio of unlabeled data and report performance in Figure \ref{fig:unlabel}. F1 performance on a fixed 10\% labeled data and 10\%, 30\%, 50\%, 70\%, 90\% unlabeled data are reported. Note that both labeled data and unlabeled data come from the training set, so we can provide unlabeled data with an upper limit of 90\%. We could see that almost all methods have performance gains with the addition of unlabeled data and {\modelname} achieves consistently better F1 performance, with a clear margin, when comparing with baselines under all different ratios of unlabeled data. 

\subsection{Analysis and Discussion}
\noindent\textbf{Effectiveness of {\gradient}}\\
The main purpose of GIRL is to guide RLG to generate pseudo labels with the similar optimization outcomes as labeled data on the unlabeled data. GIRL minimizes the discrepancy between the gradient vectors obtained from the labeled data and generated data. To demonstrate the effectiveness of {\gradient}, we first conduct an ablation study in this section. {\modelname} w/o {\gradient} is essentially the same as the Self-Training$_{BERT}$ baseline, which iteratively updates model with the synthetic set containing labeled data and generated data without {\gradient}. From Table \ref{tab:Result}, we observe {\modelname} w/o {\gradient} (Self-Training$_{BERT}$) gives us 5.38\% loss on F1, averaged over all various amounts of labeled data on two datasets. 

\begin{figure}[t!]
    \centering
    \includegraphics[width=0.49\linewidth]{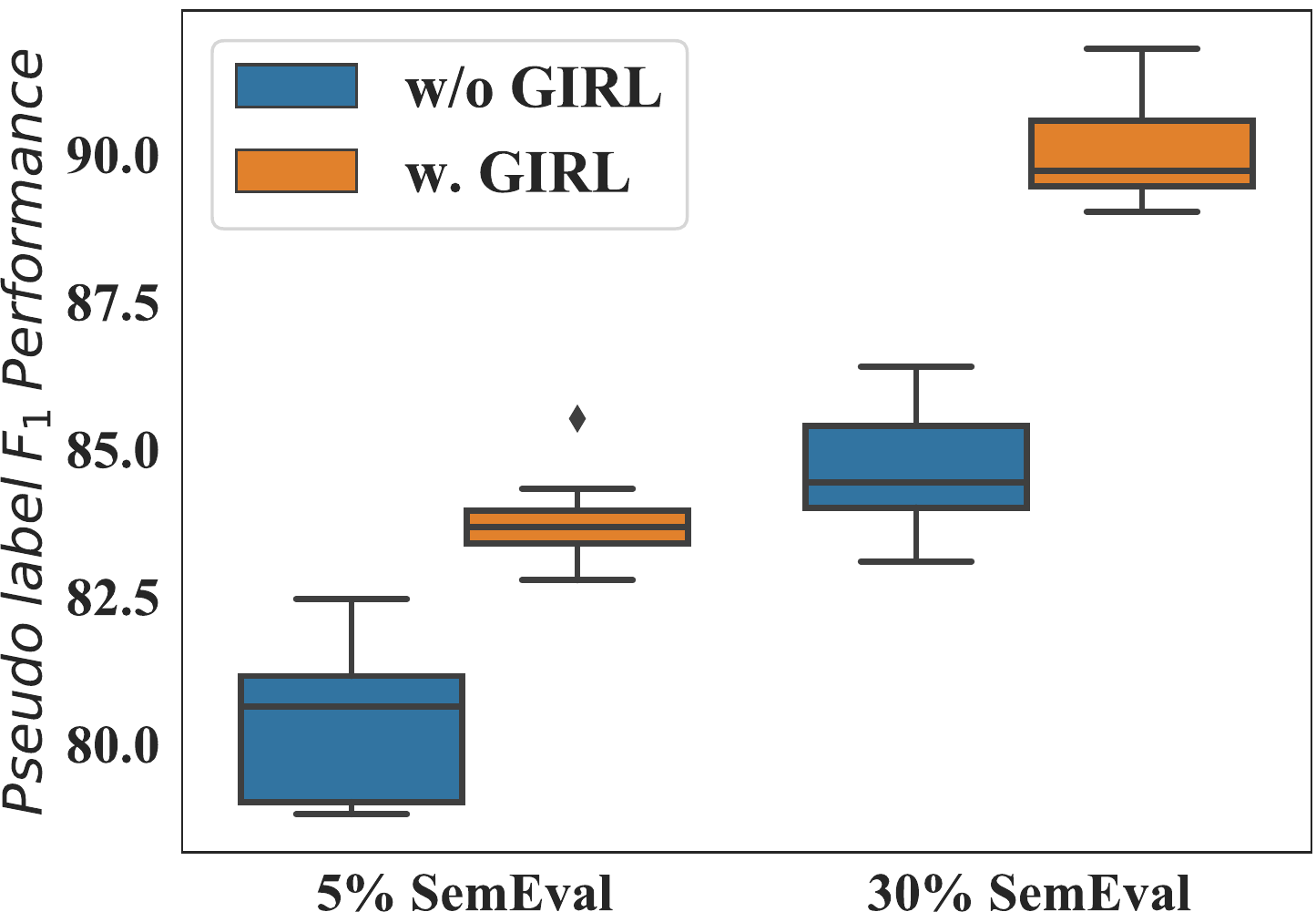} 
    \includegraphics[width=0.49\linewidth]{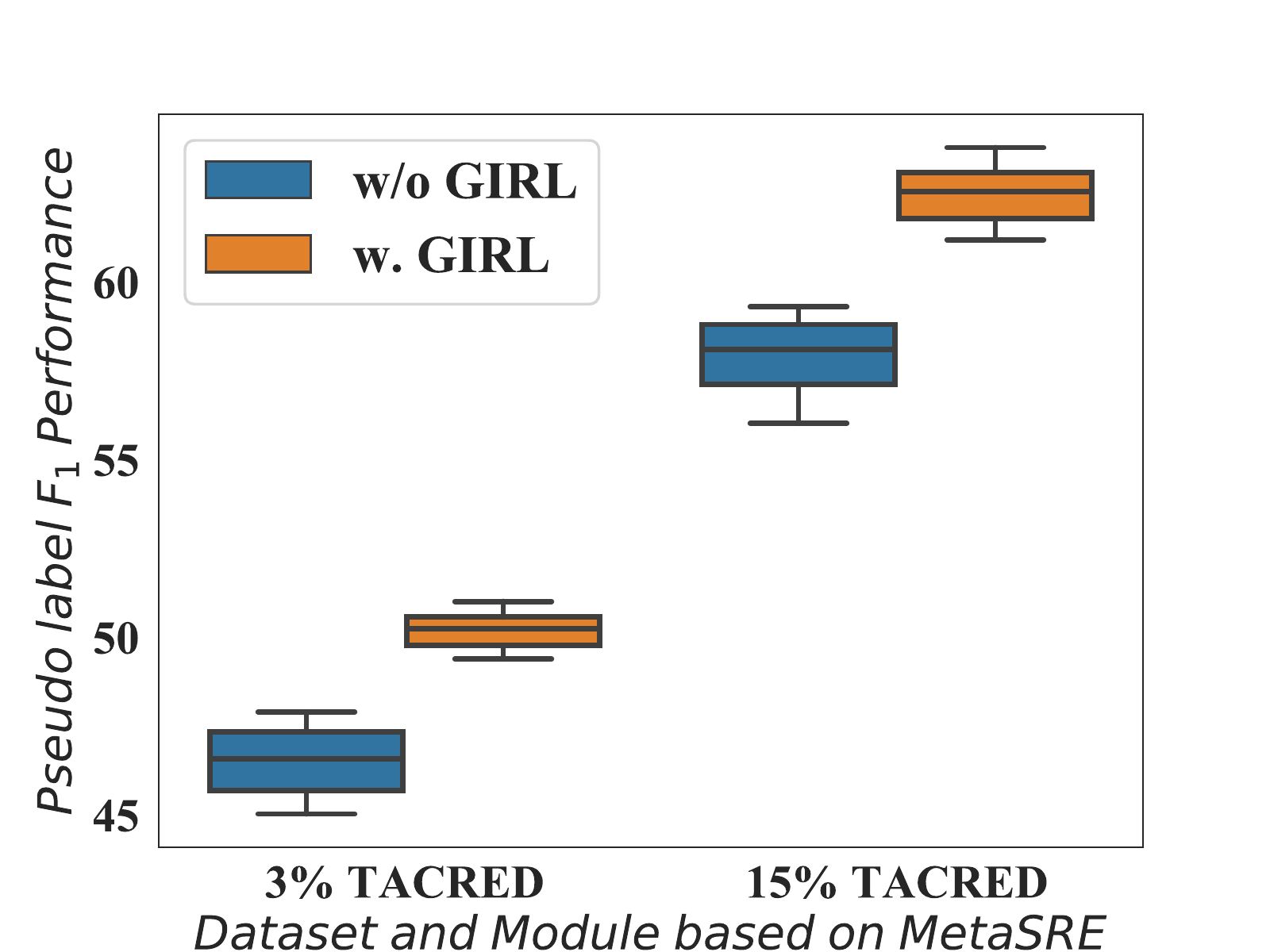}
    \caption{Pseudo label F1 (\%) Performance with GIRL based on SemEval (left) and TACRED (right).}
    \label{fig:PseudoF1}
\end{figure}

\begin{figure}[b!]
    \centering
    \includegraphics[width=0.75\linewidth]{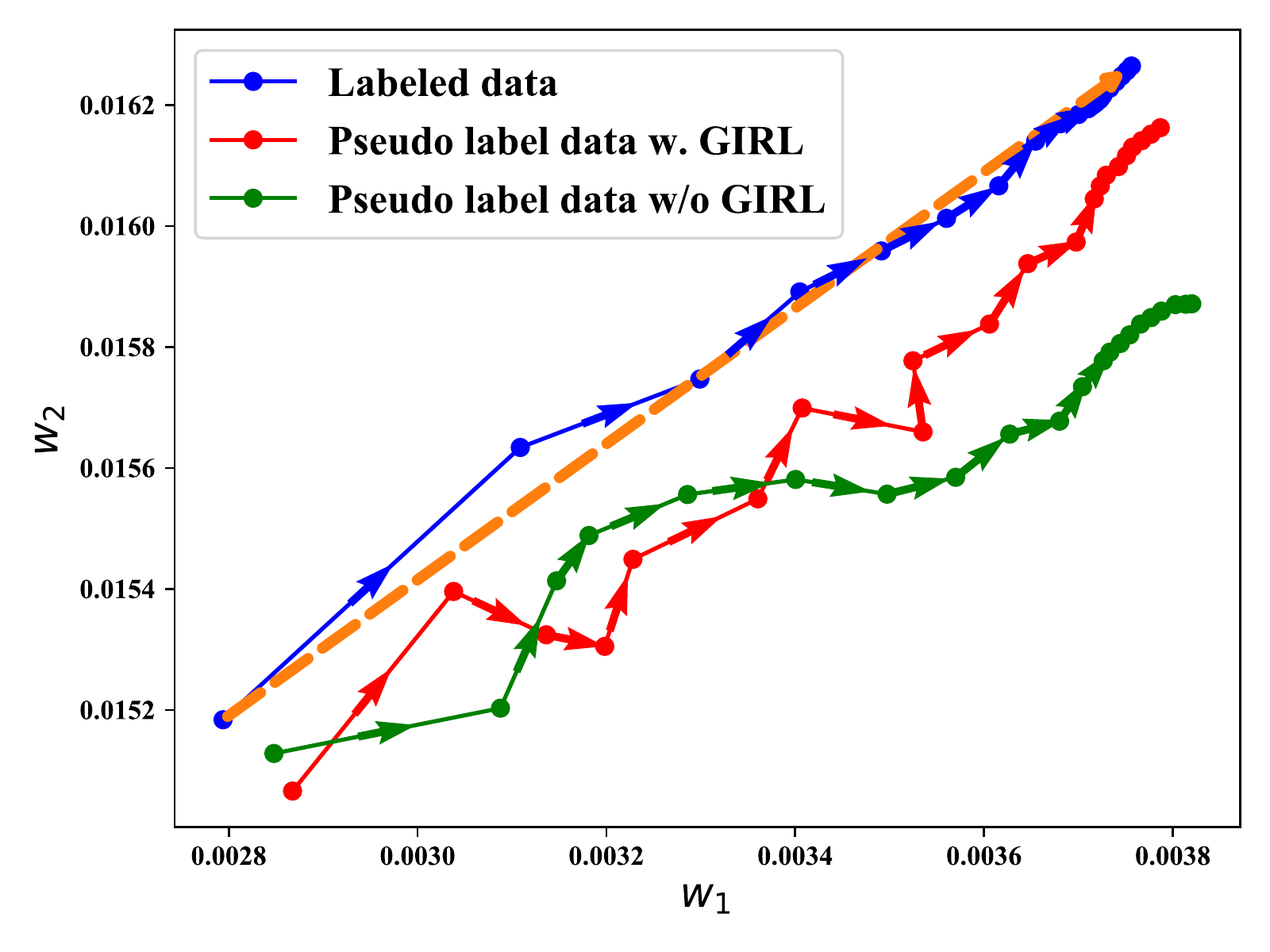}
    \caption{{\modelname} gradient descent directions on labeled data and pseudo label data. The dotted line indicates the average gradient direction on labeled data.}
    \label{fig:gradient}
\end{figure}

\begin{table}[t!]
\centering
\resizebox{0.85\linewidth}{!}{
\begin{tabular}{l}
\thickhline
\begin{tabular}[c]{@{}l@{}}My {\color{red}\textit{brother}} has entered my {\color{blue}\textit{room}} without knocking.\\ {\color{purple}Label}: \textbf{\texttt{Entity-Destination}}\\{\color{purple}Prediction w/o GIRL}: \textbf{\texttt{Other}}\\ {\color{purple}Prediction w. GIRL}: \textbf{\texttt{Entity-Destination}}\end{tabular}                                                                 \\
\\
\begin{tabular}[c]{@{}l@{}}The {\color{red}\textit{disc}} in a disc {\color{blue}\textit{music box}} plays this function,\\ with pins perpendicular to the plane surface...\\ {\color{purple}Label}: \textbf{\texttt{Content-Container}}\\{\color{purple}Prediction w/o GIRL}: \textbf{\texttt{Component-Whole}}\\ {\color{purple}Prediction w. GIRL}: \textbf{\texttt{Content-Container}}\end{tabular} \\
\\
\begin{tabular}[c]{@{}l@{}}Ditto for his funny turn as a {\color{red}\textit{man}} who instigates \\the {\color{blue}\textit{kidnapping}} of his own wife in ...\\ {\color{purple}Label}: \textbf{\texttt{Cause-Effect}}\\ {\color{purple}Prediction w/o GIRL}: \textbf{\texttt{Other}}\\ {\color{purple}Prediction w. GIRL}: \textbf{\texttt{Cause-Effect}}\end{tabular}                                       \\ \thickhline
\end{tabular}}
\caption{Predictions with/without GIRL on SemEval, where {\color{red}\textit{red}} and {\color{blue}\textit{blue}} represent head and
tail entities respectively.}\label{tab:pseudo_label_case_study}
\end{table}

We identify that the performance gains of {\modelname} come from the improved pseudo label quality by adopting GIRL.
To validate this, we draw a box plot to show the pseudo label F1. From Figure \ref{fig:PseudoF1}, we could find for the two datasets with different ratios of the labeled data, GIRL could undoubtedly improve the F1 performance of pseudo labels. In the case of 30\% SemEval and 15\% TACRED where labeled data is less scarce, GIRL can obtain more accurate gradient directions based on an increased set of labeled data.
As a result, pseudo label performance improvements are more significant.

More specifically, we show the gradient descent direction of {\modelname} on labeled data and pseudo label data in Figure \ref{fig:gradient}. Considering the overly-large parameters in RLG, we use Principal Component Analysis \cite{wold1987principal} to reduce the dimension of the parameters to $2$, and reflect the direction of gradient descent according to the update of the parameters. Although the optimization direction of pseudo label data fluctuates at the beginning, GIRL is gradually improving and ends up closer to the ideal local minima. 
When GIRL is not used, the optimization is appealing at the first because of the initial positive gains from the self-training schema. However, the error-prone pseudo labels obtained without instructive feedback gradually push the optimization away from the local minima, which leads to reduced generalization ability.

We further study cases where pseudo labels are improved with GIRL on SemEval, and present them in Table \ref{tab:pseudo_label_case_study}. {\modelname} w/o GIRL tends to predict the pseudo label as \texttt{Other} with the most occurrences, most likely because \texttt{Other} being the dominating class in the dataset. {\modelname} w. GIRL is less sensitive to the label distribution in the data and assigns correct labels. We also observe cases where GIRL is doing better at distinguishing the nuances between similar relations
such as \texttt{Content-Container} and \texttt{Component-Whole}. 

\noindent\textbf{Handling various LRE scenarios}

\begin{table}[bt!]
\centering
\resizebox{0.95\linewidth}{!}{
\begin{tabular}{lcccc}
\thickhline
&\textbf{\% Labeled Data} & \textbf{\textsf{L}} & \textbf{\textsf{L + CDA}} & \textbf{\textsf{L + U}} \\\hline
\multicolumn{1}{c}{\multirow{3}{*}{\textbf{SemEval}}} & 5\%  & 72.71                        & 75.52                     & 79.65                   \\
\multicolumn{1}{c}{}                  & 10\% & 73.93                       & 81.47                    & 81.69                   \\
\multicolumn{1}{c}{}                         & 30\% & 80.55                       & 84.63                    & {85.52}                   \\ \hline
\multicolumn{1}{c}{\multirow{3}{*}{\textbf{TACRED}}}  & 3\%  & 41.11                       & 43.34                     & {47.37}                   \\
\multicolumn{1}{c}{}                         & 10\% & 53.23                         & 57.07                     & {58.20}                    \\
\multicolumn{1}{c}{}                         & 15\% & 55.35                        & 58.89                    & {59.93}  \\\thickhline                
\end{tabular}
}
\caption{F1 (\%) of {\modelname} with various percentages of labeled data under different LRE scenarios.}\label{tab:CDA_aug}
\end{table}
\begin{figure}[bt!]
    \centering
    \includegraphics[width=0.49\linewidth]{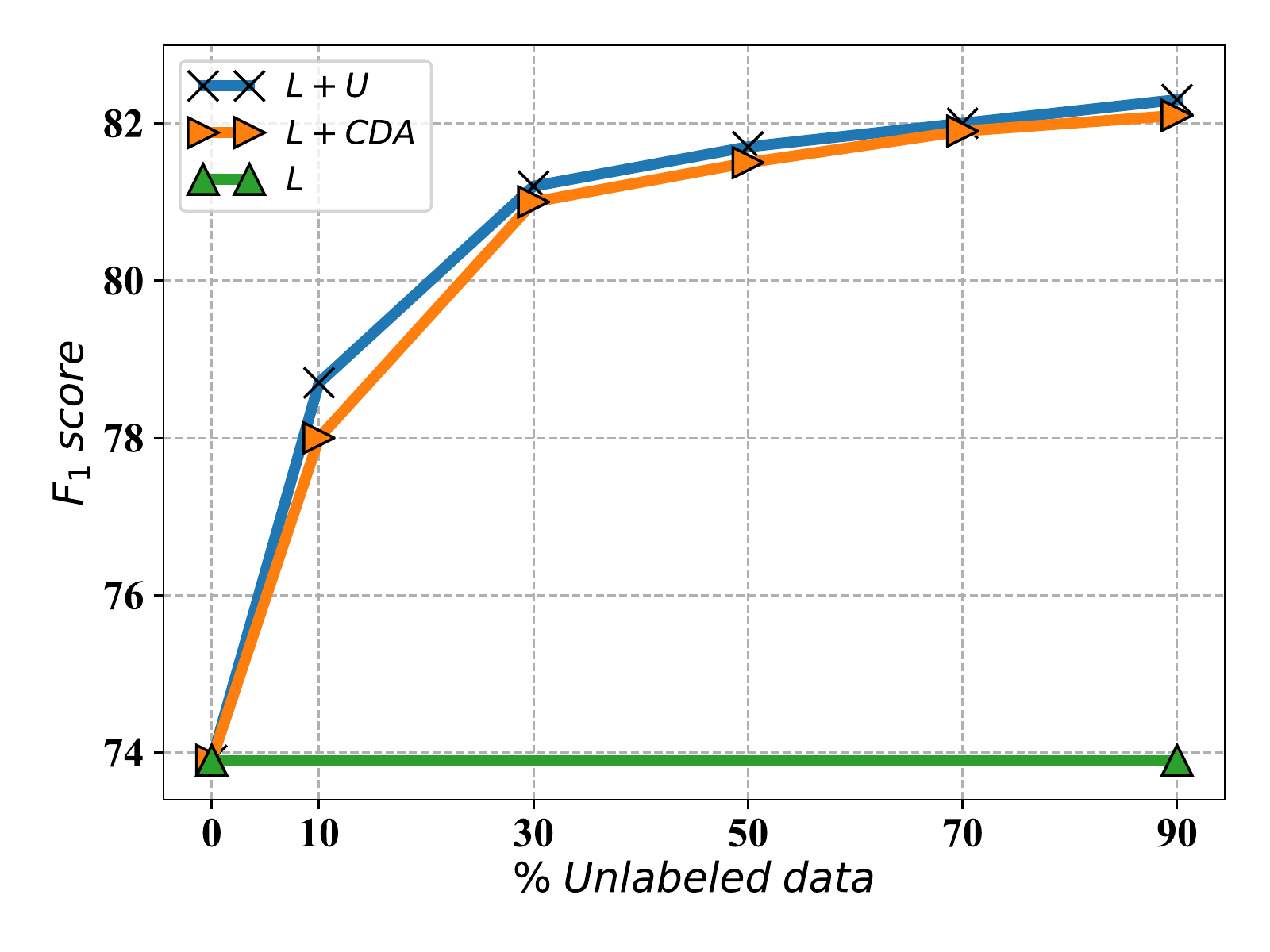}
    \includegraphics[width=0.49\linewidth]{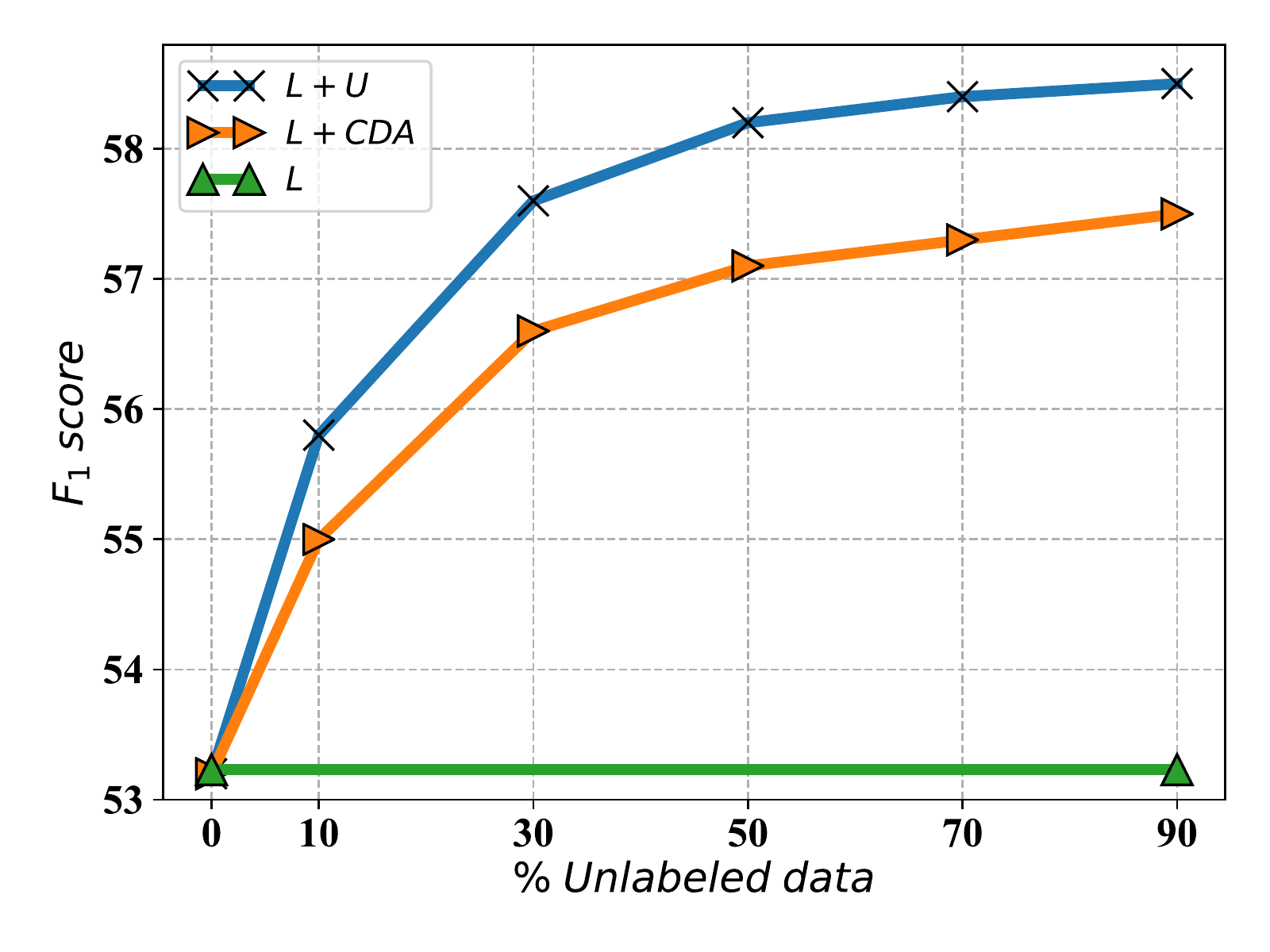}
    \caption{F1 (\%) Performance with various unlabeled data and 10\% labeled data on SemEval (left) and TACRED (right).}
    \label{fig:unlabel_aug}
\end{figure}

Considering both labeled/unlabeled data as the resource, we introduce the following LRE scenarios: 1) \textsf{L+U}: Limited labeled data and 50\% unlabeled data. 2) \textsf{L+CDA}: Only limited labeled data is available. No unlabeled data is available -- we leverage {\augmentation} (CDA) to generate the same amount of data via augmenting the labeled data. 3) \textsf{L}: This is the baseline where the model is trained only on limited labeled data. We present results in Table \ref{tab:CDA_aug}.

Compared to \textsf{L}, \textsf{L+CDA} achieves an average 4.01\% improvement in F1, indicating the effectiveness of augmentation. We also observe that \textsf{L+CDA} obtain competitive performance when compared with \textsf{L+U} on SemEval.
On a more challenging TACRED dataset, \textsf{L+CDA} achieves only 2.07\% less in F1, comparing with \textsf{L+U} when 6.36x less total samples are initially acquired. 
\begin{table}[t!]
\centering
\resizebox{0.85\linewidth}{!}{
\begin{tabular}{l}
\thickhline
\begin{tabular}[c]{@{}l@{}}Original: A {\color{red}\textit{letter}} was {\color{cyan}\textit{delivered to}} my {\color{blue}\textit{office}} in ...
\\ 
{\color{purple}Label}: \textbf{\texttt{Entity-Destination}}\\
Generated: A {\color{red}\textit{letter}} was {\color{cyan}\textit{sent from}} my {\color{blue}\textit{office}} in ...
\\{\color{purple}Pseudo label}: \textbf{\texttt{Entity-Origin}}\end{tabular}     \\
\\
\begin{tabular}[c]{@{}l@{}}Original: The {\color{red}\textit{editor}} improved the {\color{blue}\textit{manuscript}}\\ ${\quad\quad\quad\quad}$ with {\color{cyan}\textit{his changes}}.
\\ 
{\color{purple}Label}: \textbf{\texttt{Product-Producer}}\\
Generated: The {\color{red}\textit{editor}} improved the {\color{blue}\textit{manuscript}}\\ ${\quad\quad\quad\quad}$ with {\color{cyan}\textit{some improvements}}.
\\ {\color{purple}Pseudo label}: \textbf{\texttt{Product-Producer}}\end{tabular} 
\\
\\
\begin{tabular}[c]{@{}l@{}}Original: The  {\color{cyan}\textit{suspect dumped}} the {\color{cyan}\textit{dead}} {\color{red}\textit{body}}\\ ${\quad\quad\quad\quad}$  into a local {\color{blue}\textit{reservoir}}.
\\ 
{\color{purple}Label}: \textbf{\texttt{Entity-Destination}}\\
Generated: The  {\color{cyan}\textit{dam bulids}} the {\color{cyan}\textit{human}} {\color{red}\textit{body}}\\ ${\quad\quad\quad\quad}$  into a local {\color{blue}\textit{reservoir}}.
\\ {\color{purple}Pseudo label}: \textbf{\texttt{Other}}\end{tabular} \\
\thickhline
\end{tabular}
}
\caption{CDA on labeled data to obtain generated data, where {\color{red}\textit{red}} and {\color{blue}\textit{blue}} represent head and
tail entities respectively, {\color{cyan}\textit{cyan}} represents the replaced words.}\label{tab:CDA_case_study}
\vspace{-0.1in}
\end{table}

We also vary the ratio of unlabeled data (accessible by \textsf{L+U} or augmented using \textsf{L+CDA}). From Figure \ref{fig:unlabel_aug}, \textsf{L+CDA} outperforms \textsf{L} consistently, with the ratio of unlabeled data increasing, \textsf{L+CDA} can get more discriminative data and obtain better performance: it can achieve almost the same performance as \textsf{L+U} on SemEval. On TACRED, performance difference is less than 1.53\% using various ratio of unlabeled data.

We show some sample generated data produced by CDA in Table \ref{tab:CDA_case_study}. BERT Masked Language Model could generate replacement words based on the context information. 
We find that some part of the sentences with the replaced words could still maintain the original relational information, although the semantic information of another part of the sentence has changed, the RLG can still have the capability to classify the sentence into the most suitable relation.

\section{Related Work}
Relation Extraction aims to predict the binary relation between two entities in a sentence. Recent literature leverage deep neural network to encode the features among two entities from sentences, and then classify these features into pre-defined specific relation categories. These methods could gain decent performance when sufficient labeled data is available \citep{zeng2015distant,zhang2017position,DBLP:conf/ijcai/GuoN0C20,DBLP:conf/acl/NanGSL20}. However, it's labor-intensive to obtain large amounts of manual annotations on corpus.

Low resource Relation Extraction methods gained a lot of attention recently \cite{levy2017zero,tarvainen2017mean,lin2019learning,li2020exploit,hu2020semi,hu2020selfore}, since these methods require fewer labeled data and deep neural networks could expand limited labeled information by exploiting information on unlabeled data to iteratively improve the performance. One major method is the self-training work proposed by \citet{rosenberg2005semi}. Self-training incrementally assigns pseudo labels to unlabeled data and leverages these pseudo labels to iteratively improve the classification capability of the model. However, these methods always endure gradual drift problem \citep{curran2007minimising,zhang2016understanding,arazo2019unsupervised,NEURIPS2018_a19744e2,jiang2018mentornet,liu-etal-2021-noisy-labeled}: during the training process, the generated pseudo label data contains noise and could not been corrected through the model itself. Using these pseudo label data iteratively cause the model to deviate from the global minima. Our work alleviates this problem by encouraging pseudo-labeled data to imitate the gradient optimization direction on the labeled data, and introducing an effective feedback loop to improve generalization ability via reinforcement learning.

Reinforcement Learning is widely used in Nature Language Processing \cite{narasimhan2016improving,li2016deep,su2016line,yu2017seqgan,takanobu2019hierarchical}. These methods are all designed with rewards to force the correct actions to be executed during the model training process, so as to improve model performance. \citet{zeng2019learning} applies policy gradient method to model future reward in a joint entity and relation extraction task. In our work, we define reward as the cosine similarity between gradient vectors calculated from pseudo-labeled data and labeled data. 

Data augmentation methods are leveraged in natural language processing to improve the generalization ability of the model by generating discriminative samples \cite{kobayashi2018contextual,dai2020analysis,kumar2020data}. \citet{gao2019soft} contextually augment data by replacing the one-hot representation of a word by a distribution provided by BERT over the vocabulary. However, they only consider the replacement of a word which limits its capability to expand the sentence semantics\cite{joshi2020spanbert}. In our work, we use [MASK] to replace a span of words and leverage BERT Masked Language Modeling task to fill the [MASK].
\section{Conclusion}
In this paper, we propose a reinforcement learning framework model {\modelname} for low resource RE. Different from conventional self-training models which endure gradual drift when generating pseudo labels, our model encourages pseudo-labeled data to imitate the gradient optimization direction in labeled data to improve the pseudo label quality. We find our learning paradigm gives more instructive, explicit, and generalizable signals than the implicit signals that are obtained by training model directly with labeled data.
Contextualized data augmentation is proposed to handle the extremely low resource RE situation where no unlabeled data is available. Experiments on two public datasets show effectiveness of {\modelname} and augmented data over competitive baselines.
\section*{Acknowledgments}
We thank the reviewers for their valuable comments. The work was supported by the National Key Research and Development Program of China (No. 2019YFB1704003), the National Nature Science Foundation of China (No. 71690231 and No. 62021002), NSF under grants III-1763325, III-1909323, III-2106758, SaTC-1930941, Tsinghua BNRist and Beijing Key Laboratory of Industrial Bigdata System and Application.


\normalem

\bibliography{anthology,custom}
\bibliographystyle{acl_natbib}

\end{document}